\documentclass[11pt,a4paper]{article}
\usepackage[hyperref]{acl2019}
\usepackage{times}
\usepackage{latexsym}
\usepackage{graphicx}

\usepackage{url}

\aclfinalcopy % Uncomment this line for the final submission
\title{Anonymized BERT: An Augmentation Approach\\
to the Gendered Pronoun Resolution Challenge}

\author{
  Bo Liu \\
  S\&P Global\\
  New York, NY \\
  \texttt{bo.liu@spglobal.com} \\
}

\date{}

\begin{document}
\maketitle

% Abstract: four/five sentences highlighting your approach and key results.
\begin{abstract}
We present our 7th place solution\footnote{The code is available at https://github.com/boliu61/gendered-pronoun-resolution} to the Gendered Pronoun Resolution challenge, which uses BERT without fine-tuning and a novel augmentation strategy designed for contextual embedding token-level tasks. Our method anonymizes the referent by replacing candidate names with a set of common placeholder names. Besides the usual benefits of effectively increasing training data size, this approach diversifies idiosyncratic information embedded in names. Using same set of common first names can also help the model recognize names better, shorten token length, and remove gender and regional biases associated with names. The system scored 0.1947 log loss in stage 2, where the augmentation contributed to an improvements of 0.04. Post-competition analysis shows that, when using different embedding layers, the system scores 0.1799 which would be third place.
\end{abstract}

\section{Introduction}
% Introduction: ¾ a page expanding on the abstract mentioning key background such as why the task is challenging for current modeling techniques and why your approach is interesting/novel.

Gender bias has been an important topic in natural language processing in recent years \cite{bolukbasi2016man,reddy2016obfuscating,chiappa2018path,madaan2018judging}. GAP (Gendered Ambiguous Pronouns) dataset  is a gender balanced labeled corpus of 8,908 ambiguous pronoun-name pairs sampled from English Wikipedia, built and released by \newcite{webster2018gap} to challenge the community for gender unbiased pronoun resolution systems.

In the Gendered Pronoun Resolution challenge which is based on GAP dataset, we designed a unique augmentation strategy for token-level contextual embedding models and applied it to feature based BERT \cite{devlin2018bert} approach for a 7th place finish. BERT is a large bidirectional transformer trained with masked language model, which is fine-tuned to state-of-the-art results on a variety of NLP benchmark tasks. Four version of BERT model weights were released in October 2018, following a family of NLP transfer learning models in the same year, ELMo \cite{peters2018deep}, ULMFit \cite{howard2018universal} and OpenAI GPT \cite{radford2018improving}.

Although augmentation has been shown to be very effective in deep learning \cite{xie2019unsupervised}, most NLP augmentation methods are on document or sentence level, such as synonym replacement \cite{zhang2015text}, data noising \cite{xie2017data} and back-translation \cite{yu2018qanet}. For token level tasks like pronoun resolution, only the name and pronoun embeddings are in the model input. Even though altering whole document also affect these embeddings, direct change to the names has much bigger impact to the model.

The main idea of our augmentation is to replace each name in the name-pronoun pair by a set of common placeholder names, in order to (1) diversify the idiosyncratic information embedded in individual names and leave only the contextual information and (2) remove any gender or region related bias in names. In other words, to \textit{anonymize} the names and make BERT extract name-independent features purely about context. With the same set of common first names from the training corpus as the placeholders, the model can recognize candidate names more easily and embed contextual information more compactly into single tokens. This technique could also be used in other token level tasks to anonymize people or entity names.
% , such as targeted sentiment analysis \cite{jiang2011target,dong2014adaptive,vo2015target}.

\section{Model}

% Data: review of the data you used to train your system. Be sure to mention the size of the training, validation and test sets that you’ve used, and the label distributions (i.e. A, B or NEITHER), as well as any tools you used for preprocessing data (e.g. SpaCy).
% System: a detailed description of how the system was built and trained. If you’re using a neural network, were there pre-trained embeddings, how was the model trained, what hyperparameters were chosen and experimented with? How long did the model take to train, and on what infrastructure? Linking to source code is valuable here as well, but the description should be able to stand alone as a full description of how to reimplement the system. While other paper styles include background as a separate section, it’s fine to simply include citations to similar systems which inspired your work as you describe your system.

Our system is an ensemble of two neural network models, the ``End2end'' model and the ``Pure Bert'' model. 

\textbf{End2end model:} This model uses the scoring architecture proposed in \newcite{lee2017end}, but with BERT embeddings. Since candidate names A and B are already given in this task, the model doesn't have mention scores, only antecedent scores, which is a concatenation of BERT embeddings of the name (A or B); BERT embeddings of the pronoun; their element-wise similarity (between A/B and P); and non-BERT features such as distance between the name and the pronoun, whether the name is in the URL and linguistic features (syntactic distances and parts of sentence etc).

\textbf{Pure BERT model:} The input of this model is only the concatenated BERT embeddings of name A, name B and the pronoun, which are fed into two fully connected hidden layers of dimensions 512 and 32 before the softmax output layer.

\subsection{Augmentation}

Our augmentation strategy works this way: for each sample, replace all the occurrences of names A and B by 4 sets of placeholder names during both training and inference unless certain conditions are met. In training, it will make the epoch size 5 times as big. In inference, the model will make 5 predictions for each sample which are to be ensembled---this is also known as TTA (test time augmentation). 
% The TTA ensemble weights and scores are not reported due to space limit.

The 4 sets of placeholder names are
\begin{quote}
\textbf{F}: Alice, Kate, \textbf{M}: John, Michael

\textbf{F}: Elizabeth, Mary, \textbf{M}: James, Henry

\textbf{F}: Kate, Elizabeth, \textbf{M}: Michael, James

\textbf{F}: Mary, Alice, \textbf{M}: Henry, John
\end{quote}
The names were chosen from most common names in stage 1 data. For each sample, use the male pair if the pronoun is masculine (``he'', ``him'', or ``his'') and female pair otherwise.

We have experimented with fewer or more sets of placeholder names, and alternative name choices which are more ``modern'' (common names in GAP are mostly old fashioned, as many articles are about historical figures), but none worked better than the original set of names we initially chose.\\

The conditions for \textit{not} applying augmentation are:
\begin{enumerate}
\item If the placeholder name already appear in original document. e.g. in the following document, do not apply the augmentation sets that have ``Alice'' as a placeholder name,
\begin{quote}
\textbf{Alice} went to live with Nick's sister Kathy, who desperately tried to ... 
\end{quote}
\item If A or B is full name (first and last name), but the first name or last name appear alone elsewhere in the document. e.g. If we replace ``Candace Parker'' (name B) by ``Kate'' in the following sentence, the model would not known ``Kate'' and ``Parker'' are the same person
\begin{quote}
... the Shock's Plenette Pierson made a hard box-out on \textbf{Candace Parker} , causing both players to become entangled and fall over. As \textbf{Parker} tried to stand up, ...
\end{quote}
\item If the name has more than two words, such as ``Elizabeth Frances Zane'' or ``Jose de Venecia Jr'', We don't replace it because it would be difficult to implement rule 2.
\item If one of name A or B is a substring of the other, e.g. name A is ``Erin Fray'' and name B is ``Erin''. These are likely tagging errors.
\end{enumerate}

In stage 1 data, for each set of placeholder names there are 8\%, 2\%, 1\% and 1\% data that met these conditions respectively and 88\% was augmented. Note that the first 8\% are different for each set of placeholder names---only the 4\% corresponding to conditions 2-4 wasn't augmented at all.\\

\section{Experiments}

We used the official GAP dataset to build the system. There are 2000 data in both test and development sets and 454 in validation set. We used all of test and development plus 400 random rows in validation set (4400 in total) to train the system and left 54 as a sanity check to test the inference pipeline. The gender is nearly equally distributed in the training data with 2195 male and 2205 female examples.

There are 12359 samples in stage 2 test data, but only 760 were revealed to have been labeled and used for scoring. Effectively, there are 760 stage 2 test data---all the others were presumably added to prevent cheating. The gender distribution is again almost equal with 383 female and 377 male examples.

\begin{table*}[t!]
\begin{center}
\begin{tabular}{l|ll}
\hline \textbf{Model} & \textbf{End2end} & \textbf{Pure BERT} \\ \hline
ensemble weights & 0.9 & 0.1 \\
BERT embeddings & layer -4 & concatenation of layer -3 and -4\\
architecture & \newcite{lee2017end} & concatenation of A, B, Pronoun embeddings and FCN \\
non-BERT features & yes & no\\
model size & 5 MB & 36 MB\\
seed average & average of 5 seeds & only 1 seed\\
training time per seed & 30 min & 50 min\\
\hline
\end{tabular}
\caption{Meta information of two models.}\label{tab:two_models}
\end{center}
\end{table*}

The meta information for both End2end and Pure Bert model is shown in Table \ref{tab:two_models}. For each model, we trained two versions, one based on BERT Large Uncased, the other based on BERT Large Cased. For the competition, we used layer -4 (fourth to last hidden layer) embeddings for the End2end model and a concatenation of layers -3 and -4 for the Pure BERT model. As will be shown in the results section, we re-trained the models after the competition with layers -5 and -6 and achieved better results.

\textbf{Pre-processing:} As reported in the competition discussion forum, there are some clear label mistakes in GAP dataset. We identified 159 mislabels (74 development, 68 test, 17 validation) to the best of our ability by going through all the examples with a log loss of 1 or larger. We trained the system using corrected labels but report all results evaluated with original labels. 

\textbf{Post-processing:} The problem with using clean labels to train and dirty labels to evaluate is that, loss will be huge for very confident predictions if the label is wrong (i.e. when the predicted probability for the wrong-label class is very small). We solved this problem by clipping predicted probabilities smaller than a threshold 0.005, which was tuned with cross validation. The idea is similar to label smoothing \cite{szegedy2016rethinking} and confidence penalty \cite{pereyra2017regularizing}\\

All the training was done in Google Colab with a single GPU. We used 5-fold cross validation for stage 1 results, and 5-fold average for stage 2 test results. End2end model was trained 5 times using different seeds with each seed taking about 30 minutes; Pure BERT model was trained only once which took about 50 minutes.

Each team is allowed two submissions for this shared task. Above described is our submission A. Submission B is the same except that (1) it was trained on GAP test and validation sets only (2454 training samples instead of 4400), and (2) it didn't use the linguistic features. Submission B has worse results than A in both stage 1 and stage 2 as expected.

\section{Results and discussion}
% Results: a description of the key results of the paper on stage 1 and stage 2 datasets. If you have done extra error analysis into what types of errors the model makes, this is extremely valuable for the reader. Unofficial results from after the stage 2 deadline can be very useful as well.

% Discussion: general discussion of the task and your system. Description of characteristic errors and their frequency over a sample of validation data. What would you do it you had another 3 months to work on it? Are there any deployment considerations such as minimum accuracy for a useful application, fairness and bias, etc?

\begin{table*}[t!]
\begin{center}
\begin{tabular}{l|ccc}
\hline \textbf{model} & \textbf{uncased} & \textbf{cased}& \textbf{ensemble} \\ 
\hline
no augmentation &0.3878	&0.3771	&0.3470\\
augmentation only in training &0.3796	&0.3671	&0.3355\\
augmentation in both training and inference  &0.3308	&0.3308	&0.3052\\
\hline
\end{tabular}
\caption{Stage 1 results improvements in End2end model due to augmentation}\label{tab:augmentation_results}
\end{center}
\end{table*}

\subsection{Augmentation results}

In Table \ref{tab:augmentation_results}, we show the contribution of augmentation to the End2end model. In both uncased and cased versions and their ensemble, stage 1 log loss improved by about 0.01 when augmentation is added in training but not inference. And another massive 0.05 and 0.04 improvement for the uncased and cased version respectively is achieved when TTA is used. For the ensemble, augmentation improved the score from 0.3470 to 0.3052.

The reason that this augmentation method worked so well can be explained in number of ways.

1. BERT contextual embeddings of a name contain information of both the context and the name itself. Only the contextual information is relevant for coreference resolution---whether the name is Alice or Betty or Claire does not matter at all. By replacing all names by the same set of placeholders, only the useful contextual information remains for the model to learn. 

2. By using the same set of names in both training and inference, the noise from individual names are further reduced, i.e., the model will likely know they are names when it sees the same placeholder names during inference. This is even more so for foreign (non Western) names, as there are some articles in GAP about foreign figures. Without augmentation, it's less likely that BERT model trained on English corpus can recognize, for example, a lowered cased (Romanized) Chinese name as a name.

3. For gender-neutral names (including certain foreign names) and males with a typically feminine name or females with a typically masculine name, the model can much easily resolve the gender after augmentation.

4. When a long name or uncommon name is tokenized into multiple word-piece tokens, we use the average embeddings of all these tokens. Since all the placeholder names are common first names thus tokenized into single token, the syntactic information may be embedded better into a single vector than the average of a few. 

5. TTA will generate four additional predictions for each sample. Ensemble of them and the un-augmented one gives an extra boost.

Reason \#1 is related to training only, \#5 related to inference only, \#2-4 to both training and inference. An indirect proof of \#2-4 is: in TTA, the order of the 4 augmentations' scores varies depending on the model (not reported due to space limit), but they all always outperform the one without augmentation. In other words, given a trained model, the prediction on any of four augmented version is better than prediction on original data.

\subsection{Overall results}

In Table \ref{tab:model_results}, we report the log loss scores of single models and the ensemble. For stage 1, we use the 5-fold cross validation scores, trained with cleaned labels and evaluated using original labels. We also tuned the ensemble weights based on scores with cleaned labels (not shown).

During the competition, we experimented with BERT embedding layers -1 to -4 by trying different combinations of layers and their sum and concatenation and settled on layer -4 for End2end model and concatenation of -3 and -4 for Pure BERT model. After the competition ended, we realized lower layers work better on this task. So we re-trained the models using layer -5 for End2end model and layer -5 and -6 for Pure BERT model. 

The results are significantly better across the board, as shown in Table \ref{tab:model_results_rerun}. In fact, the stage 2 score 0.1799 is good enough for third place on the leaderboard. The ensemble weights were tuned on stage 1 data using clean labels as before.\\

After the competition, we also calculated the gender breakdown for all single and ensemble models based on the gender of the pronoun, reported also in Table \ref{tab:model_results} and \ref{tab:model_results_rerun}. During the competition, we trained the system and tuned the ensemble weights solely based on overall score. As a result, it exhibits some degree of gender bias in both stages, similar to \newcite{webster2018gap} and the systems cited therein. The final ensemble's bias is 0.93 in stage 1 and 0.96 in stage 2, with bias represented by the ratio of masculine and feminine scores.

Interestingly, the 4 single models demonstrate different level of bias, ranging from 0.91 to 1.03 in stage 1, and from 0.85 to 1.09 in stage 2. The larger variance is due to the much smaller stage 2 test size. Had the evaluation metrics been different than the overall log loss, we could have addressed it by assigning different weights to each single model. For instance, if systems were judged by the worse of feminine and masculine scores (to penalize heavily biased systems), we would have tuned the weights differently, sacrificing some overall score for a more balanced performance. 
For example, with ensemble weights [0.18, 0.42, 0.12, 0.28] and clipping threshold of 0.006, the overall score and gender bias of our post-competition system would be 0.2855 and 0.97 in stage 1 instead of the original version with better overall (0.2846) and a larger bias (0.93), as shown in the last row of Table \ref{tab:model_results_rerun}. On stage 2 data, the bias became slightly worse to 0.96 from 0.97. But since the stage 1 dataset is about six times as large as stage 2, the latter version is still the more gender unbiased system considering both sets.\\

\begin{figure}
\centering
\includegraphics[height=10cm]{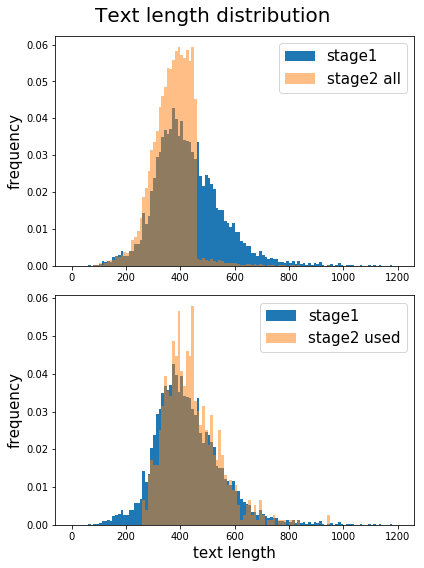}
\caption{Comparisons of document length distributions of two stages. Top: all 12359 documents in stage 2. Bottom: the 760 ``real'' documents used for scoring in stage 2.}
\label{fig:length_dist}
\end{figure}

During results checking, we noticed a clear discrepancy between the document styles of two stages. There are many more shorter documents in stage 2, as shown in the top plot of Figure \ref{fig:length_dist}. In many of the shorter documents, the pronoun refers to name A, which is the page entity. The average predicted probabilities of the three classes A, B and Neither are 0.61, 0.35 and 0.05, compared with 0.44, 0.46 and 0.10 in stage 1.

However, as revealed by the stage 2 solution, 94\% of the stage 2 data are unlabeled, which was probably generated differently (e.g. most unlabeled data have length smaller than 455). The length distribution of the 760 ``real'' labeled data used for scoring is very close to stage 1, as shown in the bottom plot of Figure \ref{fig:length_dist}. So is the predicted probability distribution (0.45, 0.46, 0.09). Then what could explain the 0.1 log loss difference between the two stage 2? We boostrapped 760 samples from stage 1 predictions for 10,000 times, the simulated stage 2 score is smaller than actual stage 2 score for only once (0.01\%). So the discrepancy is not solely due to variance from smaller sample size in stage 2. 

Our best educated guess is cleaner labels: our stage 1 score evaluated using clean labels is 0.1993, which is much closer to stage 2 score. The organizer likely spent more effort quality-checking the smaller stage 2 labels. Obviously, different pre-processing criteria during data preparation could also have made stage 2 data inherently easier to resolve.

\begin{table*}[t!]
\begin{center}
\begin{tabular}{l|c|cccc|cccc}
\hline 
&&\multicolumn{4}{c}{stage 1}&\multicolumn{4}{c}{stage 2}\\
{model} &weights&{O} & {F}&{M}&{B}&{O}&{F}&{M}&{B}\\ 
\hline
End2end, Uncased  &0.36&0.3308& 0.3439& 0.3177 &0.92& 0.2388 &0.2293&0.2484&1.08\\
End2end, Cased    &0.54&0.3308& 0.3414& 0.3201&0.94&0.2243&0.2271&0.2214&0.97\\
% End2end, Ensemble &0.3052& 0.3179& 0.2923 &0.92& 0.1938 &0.1981&0.1894&0.96\\
% \hline
Pure Bert, Uncased  &0.04& 0.3584 & 0.3649 & 0.3518 &0.96 &0.2333&0.2287&0.2381&1.04\\
Pure Bert, Cased    &0.06& 0.3544 & 0.3530 & 0.3558 &1.01 &0.2349&0.2357&0.2341&0.99\\
% Pure Bert, Ensemble & 0.3293 & 0.3302 & 0.3284 &0.99&0.2204 &0.2199&0.2209\\
\hline
Ensemble (raw)    && 0.2961 & &  &\\
Ensemble (clipped)&& \textbf{0.2922}& 0.3021&0.2823&0.93&\textbf{0.1947}&0.1983 & 0.1911&0.96\\
\hline
\end{tabular}
\caption{Log loss scores of single models and the ensemble for both stages, \textbf{competition version}, with \textbf{O}verall, \textbf{F}eminine, \textbf{M}asculine and \textbf{B}ias (M/F). Stage 2 results were evaluated after competition ended using the solution provided by Kaggle, except the final score 0.1947 (in bold), which placed 7th in the competition.}\label{tab:model_results}
\end{center}
\end{table*}

\begin{table*}[t!]
\begin{center}
\begin{tabular}{l|c|cccc|cccc}
\hline 
&&\multicolumn{4}{c}{stage 1}&\multicolumn{4}{c}{stage 2}\\
{model} &weights&{O} & {F}&{M}&{B}&{O}&{F}&{M}&{B}\\ 
\hline
End2end, Uncased   &0.36&0.3244 &0.3402 &0.3086 &0.91&0.1901&0.1820&0.1984&1.09\\
End2end, Cased     &0.44&0.3239 &0.3345 &0.3133 &0.94&0.1871&0.2017&0.1723&0.85\\
% End2end, Ensemble  &0.3006 &0.3132 &0.2879 &0.1801 &0.1856 & 0.1745\\
% \hline
Pure Bert, Uncased & 0.08&0.3486 &0.3593 &0.3378 &0.94&0.2269&0.2322&0.2215&0.95\\
Pure Bert, Cased   & 0.12&0.3492 &0.3446 &0.3539&1.03&0.2158&0.2145&0.2171&1.01\\
% Pure Bert, Ensemble& 0.3207 &0.3214 &0.3200 &0.2096 &0.2122 &0.2070\\
\hline
Ensemble (raw)&&0.2875 &&&\\
Ensemble (clipped) &&\textbf{0.2846}&0.2947&0.2744&0.93&\textbf{0.1799}&0.1829&0.1769&0.97\\
\hline
{\small More unbiased version} &&0.2855& 0.2929& 0.2780 & 0.97 & 0.1817& 0.1858& 0.1776& 0.96\\
\hline
\end{tabular}
\caption{Log loss scores of single models and the ensemble for both stages, \textbf{post-competition version}, with \textbf{O}verall, \textbf{F}eminine, \textbf{M}asculine and \textbf{B}ias (M/F). Stage 2 results were evaluated after competition ended using the solution provided by Kaggle. The stage 2 final score 0.1799 would rank third place on the leaderboard. Last row is a more gender-unbiased version with different ensemble weights.}\label{tab:model_results_rerun}
\end{center}
\end{table*}

\section{Conclusion}

% Conclusion: a restatement of the introduction, highlighting what was learned about the task and how to model it.

We presented a simple yet effective augmentation strategy that helped us finishing 7th place in the Gendered Pronoun Resolution challenge without fine-tuning. We reasoned how this technique helped the model achieving higher scores by anonymizing idiosyncrasy in individual names while also handling gender and other biases to some degree. We demonstrated how the system could be altered slightly to (1) get a better score good for 3rd place by only changing BERT embedding layers or (2) become more gender-unbiased by using different ensemble weights.

Even though our solution only used feature-based approach, we expect this augmentation method to work as well with fine-tune BERT approach, which could potentially further improve the score.

\bibliography{acl2019}

\begin{thebibliography}{16}
\expandafter\ifx\csname natexlab\endcsname\relax\def\natexlab#1{#1}\fi

\bibitem[{Bolukbasi et~al.(2016)Bolukbasi, Chang, Zou, Saligrama, and
  Kalai}]{bolukbasi2016man}
Tolga Bolukbasi, Kai-Wei Chang, James~Y Zou, Venkatesh Saligrama, and Adam~T
  Kalai. 2016.
\newblock Man is to computer programmer as woman is to homemaker? debiasing
  word embeddings.
\newblock In \emph{Advances in neural information processing systems}, pages
  4349--4357.

\bibitem[{Chiappa and Gillam(2018)}]{chiappa2018path}
Silvia Chiappa and Thomas~PS Gillam. 2018.
\newblock Path-specific counterfactual fairness.
\newblock \emph{arXiv preprint arXiv:1802.08139}.

\bibitem[{Devlin et~al.(2019)Devlin, Chang, Lee, and
  Toutanova}]{devlin2018bert}
Jacob Devlin, Ming-Wei Chang, Kenton Lee, and Kristina Toutanova. 2019.
\newblock Bert: Pre-training of deep bidirectional transformers for language
  understanding.
\newblock \emph{NAACL}.

\bibitem[{Howard and Ruder(2018)}]{howard2018universal}
Jeremy Howard and Sebastian Ruder. 2018.
\newblock Universal language model fine-tuning for text classification.
\newblock \emph{The 56th Annual Meeting of the Association for Computational
  Linguistics (ACL)}.

\bibitem[{Lee et~al.(2017)Lee, He, Lewis, and Zettlemoyer}]{lee2017end}
Kenton Lee, Luheng He, Mike Lewis, and Luke Zettlemoyer. 2017.
\newblock End-to-end neural coreference resolution.
\newblock In \emph{Proceedings of the 2017 Conference on Empirical Methods in
  Natural Language Processing}, pages 188--197.

\bibitem[{Madaan et~al.(2018)Madaan, Mehta, Mittal, and
  Suvarna}]{madaan2018judging}
Nishtha Madaan, Sameep Mehta, Shravika Mittal, and Ashima Suvarna. 2018.
\newblock Judging a book by its description: Analyzing gender stereotypes in
  the man bookers prize winning fiction.
\newblock \emph{arXiv preprint arXiv:1807.10615}.

\bibitem[{Pereyra et~al.(2017)Pereyra, Tucker, Chorowski, Kaiser, and
  Hinton}]{pereyra2017regularizing}
Gabriel Pereyra, George Tucker, Jan Chorowski, {\L}ukasz Kaiser, and Geoffrey
  Hinton. 2017.
\newblock Regularizing neural networks by penalizing confident output
  distributions.
\newblock \emph{arXiv preprint arXiv:1701.06548}.

\bibitem[{Peters et~al.(2018)Peters, Neumann, Iyyer, Gardner, Clark, Lee, and
  Zettlemoyer}]{peters2018deep}
Matthew Peters, Mark Neumann, Mohit Iyyer, Matt Gardner, Christopher Clark,
  Kenton Lee, and Luke Zettlemoyer. 2018.
\newblock Deep contextualized word representations.
\newblock In \emph{Proceedings of the 2018 Conference of the North American
  Chapter of the Association for Computational Linguistics: Human Language
  Technologies, Volume 1 (Long Papers)}, pages 2227--2237.

\bibitem[{Radford et~al.(2018)Radford, Narasimhan, Salimans, and
  Sutskever}]{radford2018improving}
Alec Radford, Karthik Narasimhan, Tim Salimans, and Ilya Sutskever. 2018.
\newblock \href
  {https://www.cs.ubc.ca/~amuham01/LING530/papers/radford2018improving.pdf}
  {Improving language understanding by generative pre-training}.

\bibitem[{Reddy and Knight(2016)}]{reddy2016obfuscating}
Sravana Reddy and Kevin Knight. 2016.
\newblock Obfuscating gender in social media writing.
\newblock In \emph{Proceedings of the First Workshop on NLP and Computational
  Social Science}, pages 17--26.

\bibitem[{Szegedy et~al.(2016)Szegedy, Vanhoucke, Ioffe, Shlens, and
  Wojna}]{szegedy2016rethinking}
Christian Szegedy, Vincent Vanhoucke, Sergey Ioffe, Jon Shlens, and Zbigniew
  Wojna. 2016.
\newblock Rethinking the inception architecture for computer vision.
\newblock In \emph{Proceedings of the IEEE conference on computer vision and
  pattern recognition}, pages 2818--2826.

\bibitem[{Webster et~al.(2018)Webster, Recasens, Axelrod, and
  Baldridge}]{webster2018gap}
Kellie Webster, Marta Recasens, Vera Axelrod, and Jason Baldridge. 2018.
\newblock Mind the gap: A balanced corpus of gendered ambiguous pronouns.
\newblock In \emph{Transactions of the ACL}, page to appear.

\bibitem[{Xie et~al.(2019)Xie, Dai, Hovy, Luong, and Le}]{xie2019unsupervised}
Qizhe Xie, Zihang Dai, Eduard Hovy, Minh-Thang Luong, and Quoc~V Le. 2019.
\newblock Unsupervised data augmentation.
\newblock \emph{arXiv preprint arXiv:1904.12848}.

\bibitem[{Xie et~al.(2017)Xie, Wang, Li, L{\'e}vy, Nie, Jurafsky, and
  Ng}]{xie2017data}
Ziang Xie, Sida~I Wang, Jiwei Li, Daniel L{\'e}vy, Aiming Nie, Dan Jurafsky,
  and Andrew~Y Ng. 2017.
\newblock Data noising as smoothing in neural network language models.
\newblock \emph{Fifth International Conference on Learning Representations
  (ICLR)}.

\bibitem[{Yu et~al.(2018)Yu, Dohan, Luong, Zhao, Chen, Norouzi, and
  Le}]{yu2018qanet}
Adams~Wei Yu, David Dohan, Minh-Thang Luong, Rui Zhao, Kai Chen, Mohammad
  Norouzi, and Quoc~V Le. 2018.
\newblock Qanet: Combining local convolution with global self-attention for
  reading comprehension.
\newblock \emph{Sixth International Conference on Learning Representations
  (ICLR)}.

\bibitem[{Zhang et~al.(2015)Zhang, Zhao, and LeCun}]{zhang2015text}
Xiang Zhang, Junbo Zhao, and Yann LeCun. 2015.
\newblock Character-level convolutional networks for text classification.
\newblock In \emph{Advances in neural information processing systems}, pages
  649--657.

\end{thebibliography}
\bibliographystyle{acl_natbib}

\end{document}